\documentclass[runningheads]{llncs}
\usepackage{graphicx}
\usepackage{hyperref,multirow, ulem,longtable}

\usepackage{amsmath}
\usepackage{amsfonts}
\usepackage{amssymb}
\usepackage{textcomp}
\usepackage{tabularx}
\usepackage{tablefootnote}

\usepackage{xcolor,todonotes}

\begin{document}
\newcommand{\significanceDef}{is denoted by **** (P$\le$1.00e-04), *** (1.00e-04$<$P$\le$1.00e-03),** (1.00e-03$<$P$\le$1.00e-02), * (1.00e-02$<$P$\le$5.00e-02), and not significant (ns) (P$>$5.00e-02)}
\newcommand{\shortsignificanceDef}{is denoted by ** (1.00e-03$<$P$\le$1.00e-02), * (1.00e-02$<$P$\le$5.00e-02), and not significant (ns) (P$>$5.00e-02)}
\newcommand{\nosignificanceDef}{is denoted by **** (P$\le$1.00e-04), *** (1.00e-04$<$P$\le$1.00e-03),** (1.00e-03$<$P$\le$1.00e-02), and * (1.00e-02$<$P$\le$5.00e-02)}
\title{A Study of Demographic Bias in CNN-based Brain MR Segmentation}
%
%
\author{Stefanos Ioannou\inst{1} \and
Hana Chockler\inst{3,1} \and
Alexander Hammers\inst{2} \and
Andrew P. King\inst{2}, for the Alzheimer’s Disease Neuroimaging
Initiative\thanks{Data used in preparation of this article were obtained from the Alzheimer’s Disease
Neuroimaging Initiative (ADNI) database (\url{adni.loni.usc.edu}). As such, the investigators
within the ADNI contributed to the design and implementation of ADNI and/or provided data
but did not participate in analysis or writing of this report. A complete listing of ADNI
investigators can be found at:
\url{http://adni.loni.usc.edu/wp-content/uploads/how_to_apply/ADNI_Acknowledgement_List.pdf}}
}

\authorrunning{S. Ioannou et al.}
%
\institute{Department of Informatics, King's College London, U.K. \and
School of Biomedical Engineering and Imaging Sciences, \\ King's College London, U.K. \and
causaLens Ltd., U.K.
}
\maketitle              
\begin{abstract}
Convolutional neural networks (CNNs) are increasingly being used to automate the segmentation of brain structures in magnetic resonance (MR) images for research studies. In other applications, CNN models have been shown to exhibit bias against certain demographic groups when they are under-represented in the training sets. In this work, we investigate whether CNN models for brain MR segmentation have the potential to contain sex or race bias when trained with imbalanced training sets. We train multiple instances of the FastSurferCNN model using different levels of sex imbalance in white subjects. We evaluate the performance of these models separately for white male and white female test sets to assess sex bias, and furthermore evaluate them on black male and black female test sets to assess potential racial bias. We find significant sex and race bias effects in segmentation model performance. The biases have a strong spatial component, with some brain regions exhibiting much stronger bias than others. Overall, our results suggest that race bias is more significant than sex bias.
Our study demonstrates the importance of considering race and sex balance when forming training sets for CNN-based brain MR segmentation, to avoid maintaining or even exacerbating existing health inequalities through biased research study findings.

\end{abstract}

\section{Introduction}

The study of bias and fairness in artificial intelligence (AI) has already attracted significant interest in the research community, with the majority of studies considering fairness in classification tasks in computer vision \cite{Mehrabi2019}. There are many causes of bias in AI, but one of the most common is the combination of imbalance and distributional shifts in the training data between protected groups\footnote{A \textit{protected group} is a set of samples which all share the same value of the \textit{protected attribute}. A protected attribute is one where fairness needs to be guaranteed, e.g. race and sex.}.
For example, \cite{Buolamwini2018a} found bias in commercial gender classification models caused by under-representation of darker-skinned people in the training set. Recently, a small number of studies have investigated bias in AI models for medical imaging applications. For example, \cite{Banerjee2021a,Larrazabal2020a} found significant under-performance on chest X-ray diagnostic models when evaluated on protected groups such as women that were under-represented in the training data. \cite{Larrazabal2020a} concluded that training set diversity and gender balance is essential for minimising bias in AI-based diagnostic decisions. Similarly, \cite{Abbasi-Sureshjani2020a} found bias in AI models for skin lesion classification and proposed a debiasing technique based on an adversarial training method. 

Whilst in computer vision classification tasks are commonplace, in medicine image segmentation plays a crucial role in many clinical workflows and research studies. For example, segmentation can be used to quantify cardiac function \cite{Ruijsink2020a} or to understand brain anatomy and development \cite{Henschel2020a}. AI techniques are increasingly being used to automate the process of medical image segmentation \cite{Isensee2021}. For example, in the brain techniques based upon convolutional neural networks (CNNs) have been proposed for automatically segmenting magnetic resonance (MR) images \cite{Coupe2020AssemblyNet:Segmentation,Henschel2020a}, outperforming the previous state of the art. However, the only study to date to have investigated bias in segmentation tasks has been \cite{Puyol-Anton2022,Puyol-Anton2021b}, which found significant racial bias in the performance of a CNN model for cardiac MR segmentation, caused by racial imbalance in the training data.

The structural anatomy of the brain is known to vary between different demographic groups, such as sex \cite{Cosgrove2007c} and race \cite{Isamah2010a}. Given that a known cause of bias in AI is the presence of such distributional shifts, and the increasing use of AI-based segmentation tools in brain imaging, it is perhaps surprising that no study to date has investigated the potential for bias in AI-based segmentation of the brain. In this paper we perform such a study. We first systematically vary levels of sex imbalance in a training set of white subjects to train multiple instances of the FastSurferCNN AI segmentation model \cite{Henschel2020a}. We evaluate the performance of these models separately using test sets comprised of white male and white female subjects to assess potential sex bias. Subsequently, we assess potential race bias by evaluating the performance of the same models on black male and black female subjects.

\section{Materials and Methods}

\subsection{Data}
To evaluate potential sex and race bias, we used MR images from the Alzheimer’s Disease Neuroimaging Initiative (ADNI)\footnote{
\url{www.adni-info.org}} database. We used a total of 715 subjects consisting of males and females of white or black race (according to the sex and race information stored in the ADNI database).

We used ground truth segmentations produced by the Multi-Atlas Label Propagation with Expectation-Maximisation based refinement (MALPEM) algorithm\footnote{We used the segmentations available at \url{https://doi.gin.g-node.org/10.12751/g-node.aa605a/}} \cite{Ledig2015a,Ledig2018}. We chose MALPEM ground truth segmentations due to the lack of manual ground truths with sufficient race/sex representation for our experiments, and the fact that MALPEM performed accurately and reliably in an extensive comparison on clinical data \cite{Johnson2017}. MALPEM segments the brain into 138 anatomical regions. See \autoref{tab:atlas} in the Supplementary Materials for a description of the MALPEM regions.

\subsection{Model and Training}

For our experiments, we used the FastSurferCNN model, which is part of the FastSurfer pipeline introduced in \cite{Henschel2020a}. The authors of the model used training data that labelled the brain following the Desikan-Killiany-Tourville (DKT) atlas, and reduced the number of anatomical regions to 78, by lateralising or combining cortical regions that are in contact with each other across hemispheres. We follow a similar approach, by lateralising, removing or combining cortical regions in the MALPEM segmentations to retain the same number of anatomical structures (i.e. 78), thus enabling us to use the FastSurferCNN model without modifying its architecture. See \autoref{tab:atlas} in the Supplementary Materials for details of how we reduced the number of anatomical regions in the MALPEM segmentations to be consistent with those expected by the FastSurferCNN model.

Models were trained using the coronal slices of the 3D MR data. The training procedure followed was identical to that of \cite{Henschel2020a}. All models were trained for 30 epochs, with a learning rate of 0.01 decreased by a factor of 0.3 every 5 epochs, a batch size of 16 and using the Adam optimiser \cite{Kingma2015}. Random translation was used to augment the training set as in
\cite{Henschel2020a}.
The loss function combines logistic loss and Dice Loss \cite{Roy2017}. A validation set of 10-20 subjects was used to monitor the training procedure. For each comparison we repeated the training twice, and report the average performance over the two runs.

The training procedure resulted in average Dice Similarity Coefficients (AVG DSC) similar to that of \cite{Henschel2020a}, therefore we assume models were trained to their full capacity.

\section{Experiments}

We now describe the experiments that we performed to investigate possible sex and race bias in the use of the FasterSurferCNN model for brain MR segmentation. We describe experiments to assess sex and race bias separately below.

\subsection{Sex Bias}
\label{sect:sexbias}

When analysing sex bias, to remove the potential confounding factor of race we used only white subjects since they had the largest number of subjects in the ADNI database.

\autoref{tab:gender-bias-runs} lists the datasets used for training/evaluation to investigate sex bias. Models were trained using training sets with different proportions of white male and white female subjects. All of the models were evaluated on the same 185 white male and 185 white female subjects. For all datasets the scanner manufacturer (Siemens, GE Medical Systems, and Philips), subject age, field strength (3.0T and 1.5T), and diagnosis (Dementia, Mild Cognitive Impairment, and Cognitively Normal) were controlled for. All images were acquired using the MP-RAGE sequence, which showed the highest performance in \cite{Henschel2020a}.

\begin{table}
\centering
\begin{tabular}{|l|l|l|l|l|}
\hline
\textbf{Usage} & \textbf{Female, n (\%)\ } & \textbf{Male, n (\%)\ } & \textbf{Age $\pm$ SD\ } & \textbf{Description}\\ \hline
\multirow{5}{*}{Training} & 140 (100) & 0 (0) & 74.29 $\pm$ 5.52 & 100\% female\\ \cline{2-5} 
 & 35 (25) & 105 (75) & 75.6 $\pm$ 4.82 & 75\% male, 25\% female\\ \cline{2-5} 
 & 70 (50) & 70 (50) & 75.13 $\pm$ 5.34 & 50\% male, 50\% female\\ \cline{2-5} 
 & 105 (75) & 35 (25) & 74.64 $\pm$ 5.64 & 25\% male, 75\% female\\ \cline{2-5} 
 & 0 (0) & 140 (100) & 75.68 $\pm$ 4.84 & 100\% male\\ \hline
\multirow{2}{*}{Testing} & 185 (100) & 0 (0) & 74.44 $\pm$ 5.16 & Female test set\\ \cline{2-5} 
 & 0 & 185 (100) & 75.55 $\pm$ 5.55 & Male test set\\ \hline
\end{tabular}
\caption{Training and test sets used to assess sex bias. The number (and proportion) of white female and white male subjects and the mean and standard deviations (SD) of the subjects' ages in each dataset.}
\label{tab:gender-bias-runs}
\end{table}

\subsection{Racial Bias}
\label{sect:racebias}

To investigate racial bias, the same models trained with white subjects in the sex bias experiment (see \autoref{tab:gender-bias-runs}) were evaluated on test sets broken down by both sex and race, utilising an extra set of male/female black race subjects. We used four test sets in total in this experiment: white female, white male, black female and black male. All test sets consisted of 36 subjects each and were controlled for age, scanner manufacturer, field strength and diagnosis as in the sex bias experiment.

\subsection{Evaluation and Statistical Analysis}
To evaluate the performance of the models, we computed the DSC on a per-region basis as well as the generalised DSC (GDSC) \cite{Sudre2017} (excluding the background class) to quantify overall performance. All metrics were computed for each test set individually as detailed in Sections \ref{sect:sexbias} and \ref{sect:racebias}. Statistical tests were performed using Wilcoxon signed rank tests (for paired comparisons) and Mann-Whitney U tests (for unpaired comparisons), both using $(p \le 0.01)$ significance between DSC values for different models.\footnote{When assessing differences for multiple regions we did not apply correction for multiple tests because our aim was to be sensitive to possible bias rather than minimise Type I errors.}

\section{Results}
\subsection{Sex Bias}

The AVG GDSC results (see \autoref{fig:gender-bias-overall} in the Supplementary Materials) showed that the models exhibit some signs of potential bias for both white males and white females. However, most of the comparisons showed no statistical significance when considering the GDSC overall. Therefore, we more closely analysed individual brain regions that did show statistically significant differences in performance in terms of per-region AVG DSC.

\begin{figure}
    \centering
    \includegraphics[width=300pt]{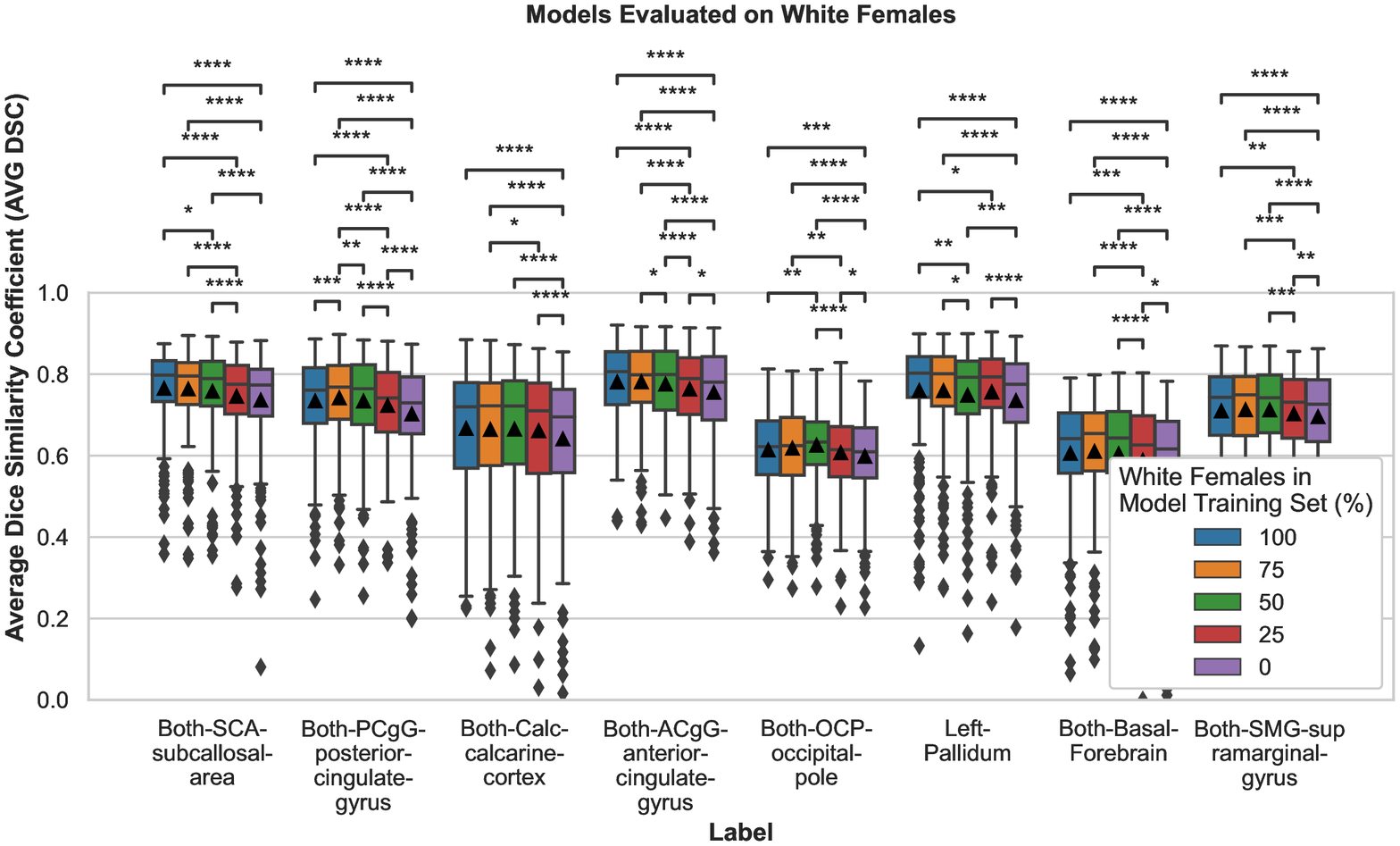}
    \caption{Regions showing the highest bias according to models' DSC performance on white females. Significance using Wilcoxon signed ranked test \nosignificanceDef. }
    \label{fig:gender-bias-test-female}
\end{figure}

\begin{figure}
    \centering
    \includegraphics[width=300pt]{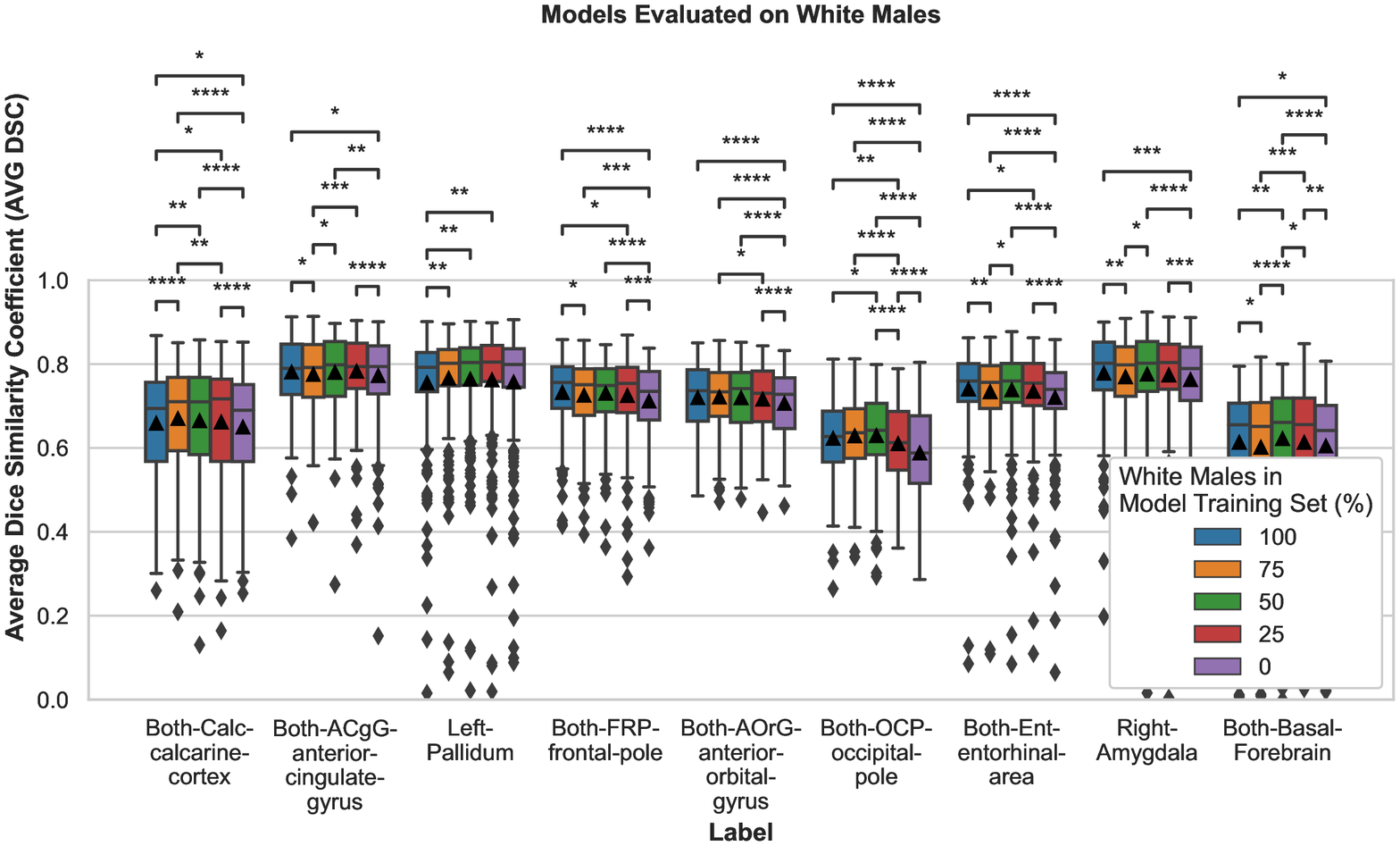}
    \caption{Regions showing the highest bias according to models' DSC performance on white males. Significance using Wilcoxon signed ranked test \nosignificanceDef.}
    \label{fig:gender-bias-test-male}
\end{figure}

\autoref{fig:gender-bias-test-female} shows the performance of the models for selected regions exhibiting the highest bias when evaluated on white female subjects. As can be seen, AVG DSC for these regions decreases as the proportion of female subjects in the training set decreases. Over all regions, when evaluating on white females the model trained using 100\% male subjects had significantly lower AVG DSC compared to the model trained with 100\% females in 53 of the 78 regions $(p \le 0.01)$. The 	Both-PCgG-posterior-cingulate-gyrus region exhibits the highest decrease in AVG-DSC of 0.0395.
Similarly, when evaluating on white males (\autoref{fig:gender-bias-test-male}), the AVG DSC for regions showing the highest bias decreases with the proportion of males in the training set. This time, 36 of the 78 regions had significantly worse performance for the 100\% female trained model compared to the 100\% male trained model $(p \le 0.01)$. The Both-OCP-occipital-pole region exhibits the highest decrease in AVG-DSC of 0.0406. For both the white male and white female test sets, no region showed statistical significance in the opposite direction. Interestingly, we note that for the majority of regions, all models perform better on males than on females, even when the model is trained only on female subjects. For instance, the model trained with a sex-balanced dataset shows a significant decrease ($p \le 0.01$ using the Mann-Whitney U test) in 3 regions when evaluated on white female subjects compared to white male subjects, with no significant difference in the opposite direction. The Both-Inf-Lat-Vent region shows the highest decrease in AVG-DSC of 0.0475.

\subsection{Racial Bias}

For assessing racial bias, the AVG GDSC results (see  \autoref{fig:race-bias-overall} in the Supplementary Materials) show that the models again exhibit some signs of bias but most comparisons are not statistically significant. However, it can be observed that statistically significant differences are found when using a black female test set.

\begin{figure}
    \centering
    \includegraphics[width=300pt]{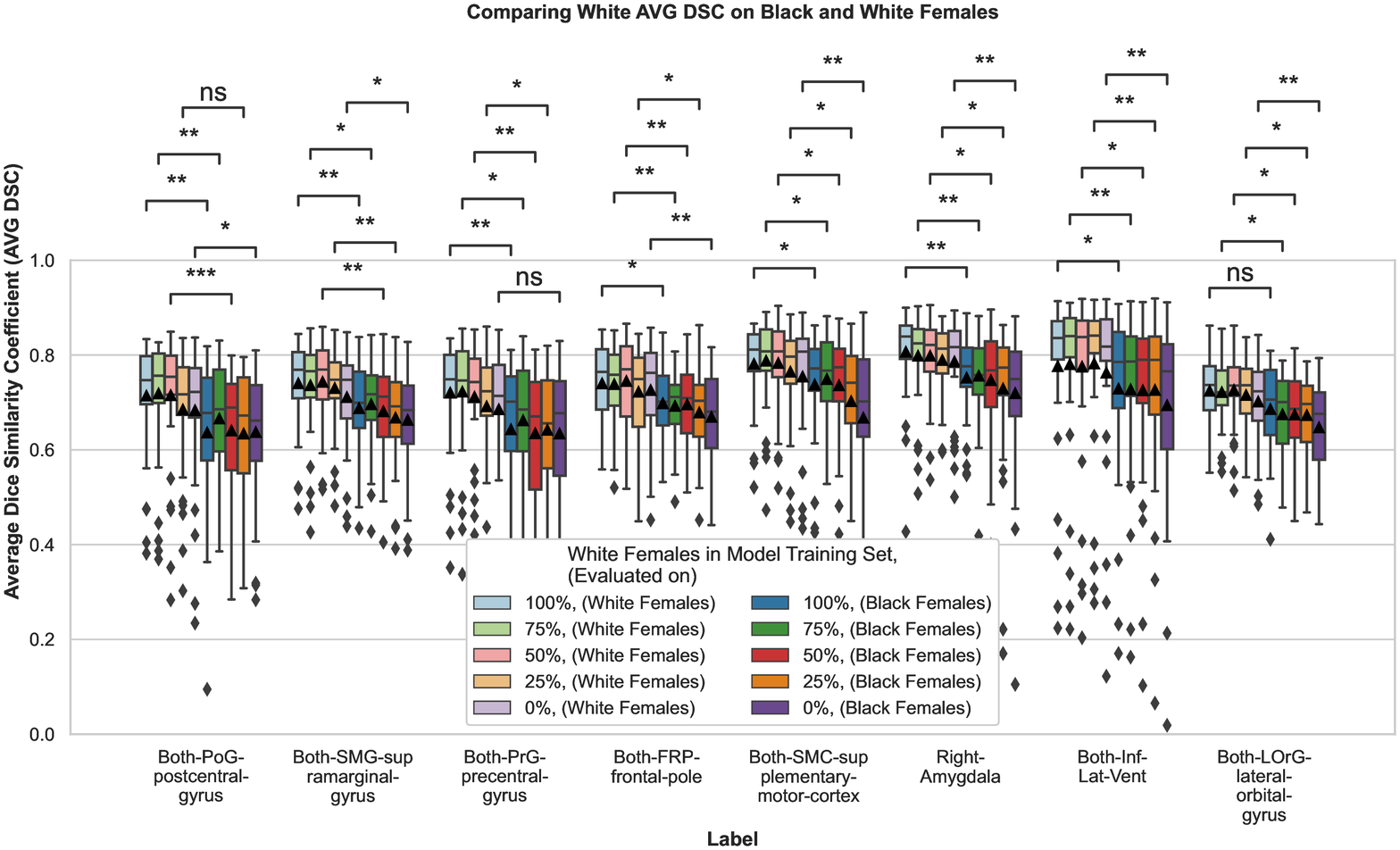}
    \caption{Comparing AVG DSC performance of models trained with white subjects, evaluated on white and black female subjects. Mann-Whitney U test \significanceDef.}
    \label{fig:race-bias-female}
\end{figure}

As in the sex bias experiments, we more closely analysed individual regions that showed the highest bias in AVG DSC. \autoref{fig:race-bias-female} summarises these results for white and black female test sets. We can see a significant drop in performance when the (white-trained) models are evaluated on black female subjects. 
This difference in performance becomes more pronounced as the proportion of female subjects decreases, indicating a possible interaction between sex and race bias.
When comparing test performance on black female subjects compared to white female subjects, the model trained only on (white) female subjects exhibits the highest decrease in AVG-DSC of 0.0779 in the 	Both-PoG-postcentral-gyrus region.
The model trained only on (white) males shows the highest decrease in AVG-DSC of 0.0868 in the Both-SMC-supplementary-motor-cortex region. No region showed statistically significantly higher AVG DSC in black subjects compared to white, for either sex.

\begin{figure}
    \centering
    \includegraphics[width=250pt]{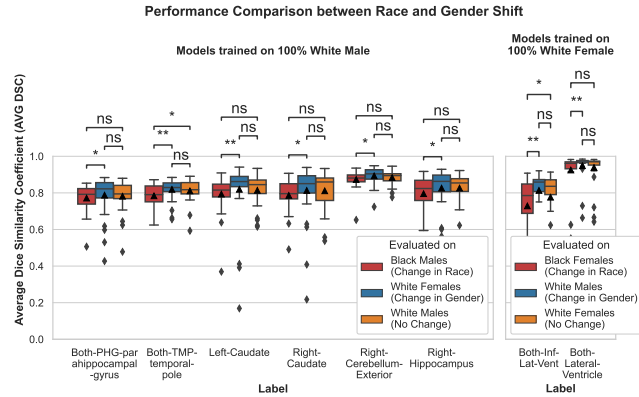}
    \caption{Comparison of gender and race bias. The plots include all labels with a statistically significant change in performance by sex or race, of models trained on white males (left) and white females (right). Significance using Mann-Whitney U test \shortsignificanceDef.}
    \label{fig:race-vs-gender}
\end{figure}

We also performed a comparison (see \autoref{fig:race-vs-gender}) between the effects of sex and race bias and found that the race bias effect was more significant than the sex bias effect.

\section{Discussion}

To the best of our knowledge, this paper has presented the first study of demographic bias in CNN-based brain MR segmentation. The study is timely because CNN models are increasingly being used to derive biomarkers of brain anatomy for clinical research studies. If certain demographic groups suffer worse performance in such models this will lead to bias in the findings of the research studies, leading to the maintaining or even exacerbation of existing health inequalities.

Our study found that CNNs for brain MR segmentation can exhibit significant sex and race bias when trained with imbalanced training sets. This is likely due to the algorithm introducing bias and/or the well-known effect of representation bias \cite{Mehrabi2019b}, in which distributional shifts combined with data imbalance lead to biased model performance. Interestingly, the biases we found have a strong spatial component, with certain brain regions exhibiting a very pronounced bias effect (in both sex and race), whilst others show little or no difference. This is likely caused by a similar spatial component in the distributional shift, i.e. differences in brain anatomy between sexes and races are likely to be localised to certain regions. We found that sex bias in performance still exists even when the model's training set is sex balanced. This is likely due to algorithmic rather than representation bias.
Overall, we found that the effect of race bias was stronger than that of sex bias. Furthermore, the bias effect was much more pronounced in black females than black males.

We also observed that there was not always a monotonically increasing/ decreasing trend in the performance of the models as we changed the level of imbalance. In particular, it was often the case that models trained with a (small) proportion of a different protected group improved performance for the majority group. We speculate that including data from a different protected group(s) can increase diversity in the training set, hence improving the generalisation ability of the model and leading to better performance for all protected groups.

We believe that our findings are important for the future use of CNNs in clinical research based on neuroimaging. However, we acknowledge a number of limitations. First, the number of subjects we could employ in the study was necessarily limited. The majority of subjects in the ADNI database with available MALPEM ground truth segmentations are Caucasian (i.e. white) and so we were limited in the numbers of non-white subjects we could make use of. This prevented us from performing a systematic study of the impact of race imbalance, similar to the way in which we trained multiple models with varying sex imbalance in the sex bias experiment. Another limitation is that we could not employ manual ground truth segmentations in our study. Again, this was because of the lack of large numbers of manual ground truth segmentations that are publicly available, particularly for non-white races. Although the MALPEM segmentations used in this study have been quality-inspected, we cannot assume that they are unbiased according to race and sex, or free of other systematic and random errors. Even so, we believe that the disparities in performance observed in our study can be regarded as model-induced bias, perhaps in addition to that which might be present in the ground-truth segmentations.

Another limitation is that, in this work, we only used the coronal orientation of the 3-D MR images for training and evaluation of the models. Training a model using all orientations will be the subject of future work. Future work will also focus on more extensive and detailed studies of demographic bias in brain MR segmentation, as well as investigation of techniques for bias mitigation \cite{Mehrabi2019b}.

%
%
%
\bibliographystyle{splncs04}
\bibliography{references}

\begin{thebibliography}{10}
\providecommand{\url}[1]{\texttt{#1}}
\providecommand{\urlprefix}{URL }
\providecommand{\doi}[1]{https://doi.org/#1}

\bibitem{Abbasi-Sureshjani2020a}
Abbasi-Sureshjani, S., Raumanns, R., Michels, B.E., Schouten, G., Cheplygina,
  V.: {Risk of Training Diagnostic Algorithms on Data with Demographic Bias}.
  Lecture Notes in Computer Science (including subseries Lecture Notes in
  Artificial Intelligence and Lecture Notes in Bioinformatics)  \textbf{12446
  LNCS},  183--192 (2020). \doi{10.1007/978-3-030-61166-8{\_}20}

\bibitem{Banerjee2021a}
Banerjee, I., Bhimireddy, A.R., Burns, J.L., Celi, L.A., Chen, L.C., Correa,
  R., Dullerud, N., Ghassemi, M., Huang, S.C., Kuo, P.C., Lungren, M.P.,
  Palmer, L., Price, B.J., Purkayastha, S., Pyrros, A., Oakden-Rayner, L.,
  Okechukwu, C., Seyyed-Kalantari, L., Trivedi, H., Wang, R., Zaiman, Z.,
  Zhang, H., Gichoya, J.W.: {Reading Race: AI Recognises Patient's Racial
  Identity In Medical Images}  (2021)

\bibitem{Buolamwini2018a}
Buolamwini, J.: {Gender Shades: Intersectional Accuracy Disparities in
  Commercial Gender Classification *}. Proceedings of Machine Learning Research
   \textbf{81},  1--15 (2018)

\bibitem{Cosgrove2007c}
Cosgrove, K.P., Mazure, C.M., Staley, J.K.: {Evolving Knowledge of Sex
  Differences in Brain Structure, Function, and Chemistry}. Biological
  Psychiatry  \textbf{62}(8),  847--855 (2007).
  \doi{10.1016/j.biopsych.2007.03.001}

\bibitem{Coupe2020AssemblyNet:Segmentation}
Coup{\'{e}}, P., Mansencal, B., Cl{\'{e}}ment, M., Giraud, R., Denis~de
  Senneville, B., Ta, V.T., Lepetit, V., Manjon, J.V.: {AssemblyNet: A large
  ensemble of CNNs for 3D whole brain MRI segmentation}. NeuroImage
  \textbf{219},  117026 (10 2020). \doi{10.1016/J.NEUROIMAGE.2020.117026}

\bibitem{Henschel2020a}
Henschel, L., Conjeti, S., Estrada, S., Diers, K., Fischl, B., Reuter, M.:
  {FastSurfer - A fast and accurate deep learning based neuroimaging pipeline}.
  NeuroImage  \textbf{219},  117012 (10 2020).
  \doi{10.1016/j.neuroimage.2020.117012}

\bibitem{Isamah2010a}
Isamah, N., Faison, W., Payne, M.E., MacFall, J., Steffens, D.C., Beyer, J.L.,
  Krishnan, K.R., Taylor, W.D.: {Variability in frontotemporal brain structure:
  The importance of recruitment of African Americans in neuroscience research}.
  PLoS ONE  \textbf{5}(10),  e13642 (2010). \doi{10.1371/journal.pone.0013642}

\bibitem{Isensee2021}
Isensee, F., Jaeger, P.F., Kohl, S.A., Petersen, J., Maier-Hein, K.H.:
  {nnU-Net: a self-configuring method for deep learning-based biomedical image
  segmentation}. Nature Methods  \textbf{18}(2),  203--211 (12 2021).
  \doi{10.1038/s41592-020-01008-z}

\bibitem{Johnson2017}
Johnson, E.B., Gregory, S., Johnson, H.J., Durr, A., Leavitt, B.R., Roos, R.A.,
  Rees, G., Tabrizi, S.J., Scahill, R.I.: {Recommendations for the use of
  automated gray matter segmentation tools: Evidence from Huntington's
  disease}. Frontiers in Neurology  \textbf{8}(OCT), ~519 (10 2017).
  \doi{10.3389/fneur.2017.00519}

\bibitem{Kingma2015}
Kingma, D.P., Ba, J.L.: {Adam: A method for stochastic optimization}. 3rd
  International Conference on Learning Representations, ICLR 2015 - Conference
  Track Proceedings  (12 2015). \doi{10.48550/arxiv.1412.6980}

\bibitem{Larrazabal2020a}
Larrazabal, A.J., Nieto, N., Peterson, V., Milone, D.H., Ferrante, E.: {Gender
  imbalance in medical imaging datasets produces biased classifiers for
  computer-aided diagnosis}. Proceedings of the National Academy of Sciences of
  the United States of America  \textbf{117}(23),  12592--12594 (6 2020).
  \doi{10.1073/pnas.1919012117}

\bibitem{Ledig2015a}
Ledig, C., Heckemann, R.A., Hammers, A., Lopez, J.C., Newcombe, V.F.,
  Makropoulos, A., L{\"{o}}tj{\"{o}}nen, J., Menon, D.K., Rueckert, D.: {Robust
  whole-brain segmentation: Application to traumatic brain injury}. Medical
  Image Analysis  \textbf{21}(1),  40--58 (4 2015).
  \doi{10.1016/j.media.2014.12.003}

\bibitem{Ledig2018}
Ledig, C., Schuh, A., Guerrero, R., Heckemann, R.A., Rueckert, D.: {Structural
  brain imaging in Alzheimer’s disease and mild cognitive impairment:
  biomarker analysis and shared morphometry database}. Scientific Reports
  \textbf{8}(1),  11258 (2018). \doi{10.1038/s41598-018-29295-9}

\bibitem{Mehrabi2019b}
Mehrabi, N., Morstatter, F., Saxena, N., Lerman, K., Galstyan, A.: {A Survey on
  Bias and Fairness in Machine Learning}. ACM Computing Surveys  \textbf{54}(6)
  (8 2019). \doi{10.1145/3457607}

\bibitem{Mehrabi2019}
Mehrabi, N., Morstatter, F., Saxena, N., Lerman, K., Galstyan, A.: {A Survey on
  Bias and Fairness in Machine Learning}. ACM Computing Surveys  \textbf{54}(6)
  (8 2021). \doi{10.1145/3457607}

\bibitem{Puyol-Anton2022}
Puyol-Ant{\'{o}}n, E., Ruijsink, B., Mariscal~Harana, J., Piechnik, S.K.,
  Neubauer, S., Petersen, S.E., Razavi, R., Chowienczyk, P., King, A.P.:
  {Fairness in Cardiac Magnetic Resonance Imaging: Assessing Sex and Racial
  Bias in Deep Learning-Based Segmentation}. Frontiers in Cardiovascular
  Medicine  \textbf{9}, ~664 (4 2022). \doi{10.3389/fcvm.2022.859310}

\bibitem{Puyol-Anton2021b}
Puyol-Ant{\'{o}}n, E., Ruijsink, B., Piechnik, S.K., Neubauer, S., Petersen,
  S.E., Razavi, R., King, A.P.: {Fairness in Cardiac MR Image Analysis: An
  Investigation of Bias Due to Data Imbalance in Deep Learning Based
  Segmentation}. Lecture Notes in Computer Science (including subseries Lecture
  Notes in Artificial Intelligence and Lecture Notes in Bioinformatics)
  \textbf{12903 LNCS},  413--423 (2021). \doi{10.1007/978-3-030-87199-4{\_}39}

\bibitem{Roy2017}
Roy, A.G., Conjeti, S., Sheet, D., Katouzian, A., Navab, N., Wachinger, C.:
  {Error corrective boosting for learning fully convolutional networks with
  limited data}. Lecture Notes in Computer Science (including subseries Lecture
  Notes in Artificial Intelligence and Lecture Notes in Bioinformatics)
  \textbf{10435 LNCS},  231--239 (9 2017).
  \doi{10.1007/978-3-319-66179-7{\_}27}

\bibitem{Ruijsink2020a}
Ruijsink, B., Puyol-Ant{\'{o}}n, E., Oksuz, I., Sinclair, M., Bai, W.,
  Schnabel, J.A., Razavi, R., King, A.P.: {Fully Automated, Quality-Controlled
  Cardiac Analysis From CMR: Validation and Large-Scale Application to
  Characterize Cardiac Function}. JACC: Cardiovascular Imaging  \textbf{13}(3),
   684--695 (3 2020). \doi{10.1016/j.jcmg.2019.05.030}

\bibitem{Sudre2017}
Sudre, C.H., Li, W., Vercauteren, T., Ourselin, S., Jorge~Cardoso, M.:
  {Generalised Dice overlap as a deep learning loss function for highly
  unbalanced segmentations}. Lecture Notes in Computer Science (including
  subseries Lecture Notes in Artificial Intelligence and Lecture Notes in
  Bioinformatics)  \textbf{10553 LNCS},  240--248 (2017).
  \doi{10.1007/978-3-319-67558-9{\_}28}

\end{thebibliography}


\newpage
\section*{Supplementary Material}
\setcounter{figure}{0}
\renewcommand{\thetable}{S\arabic{table}}
\renewcommand{\thefigure}{S\arabic{figure}}

 \begin{figure}
     \centering
     \includegraphics[height=128pt]{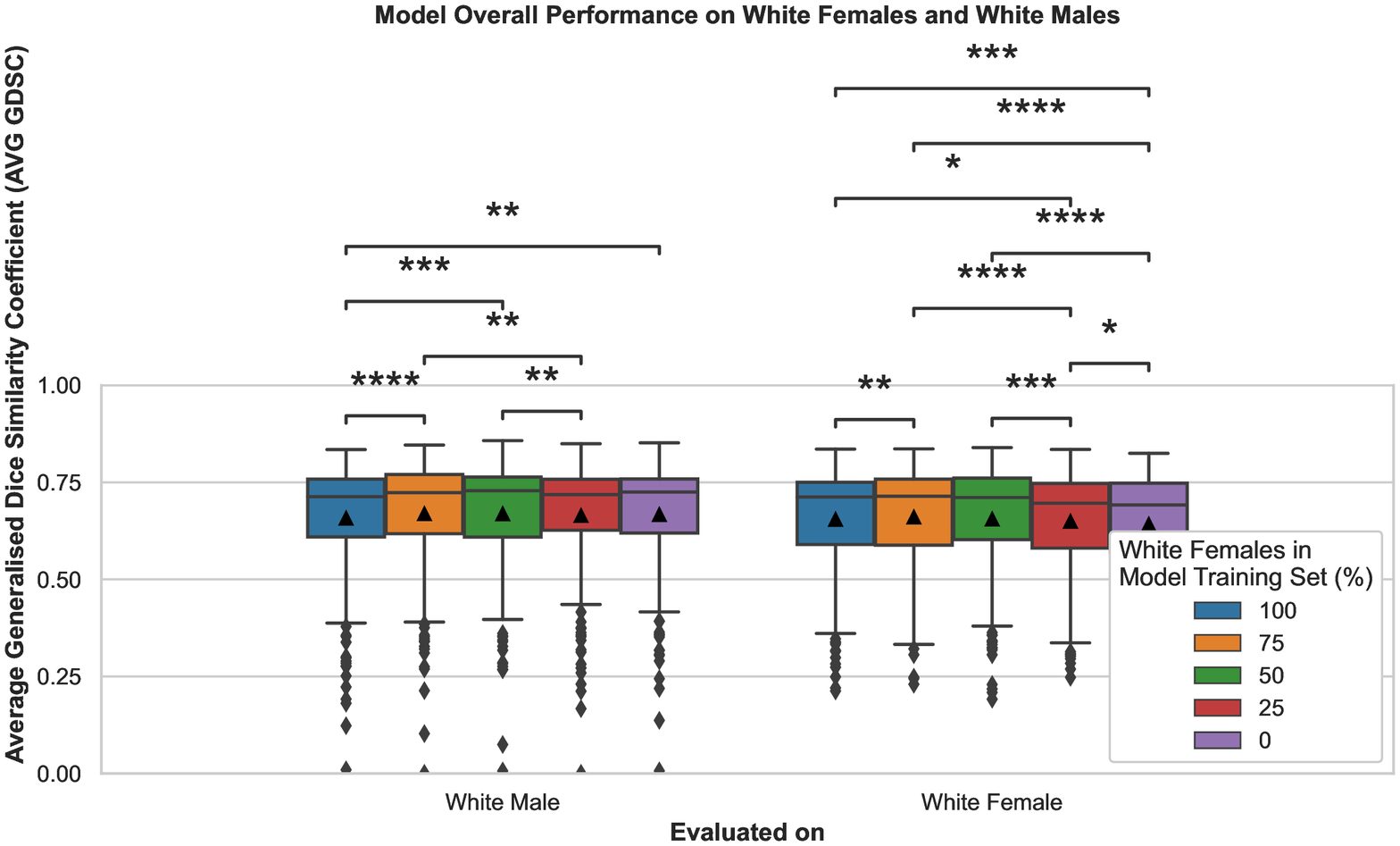}
     \caption{AVG GDSC (excluding background class) on white females and white males using different levels of sex imbalance in the training set. Significance using Wilcoxon signed rank test \nosignificanceDef.}
     \label{fig:gender-bias-overall}
 \end{figure} 

 \begin{figure}
     \centering
     \includegraphics[height=130pt]{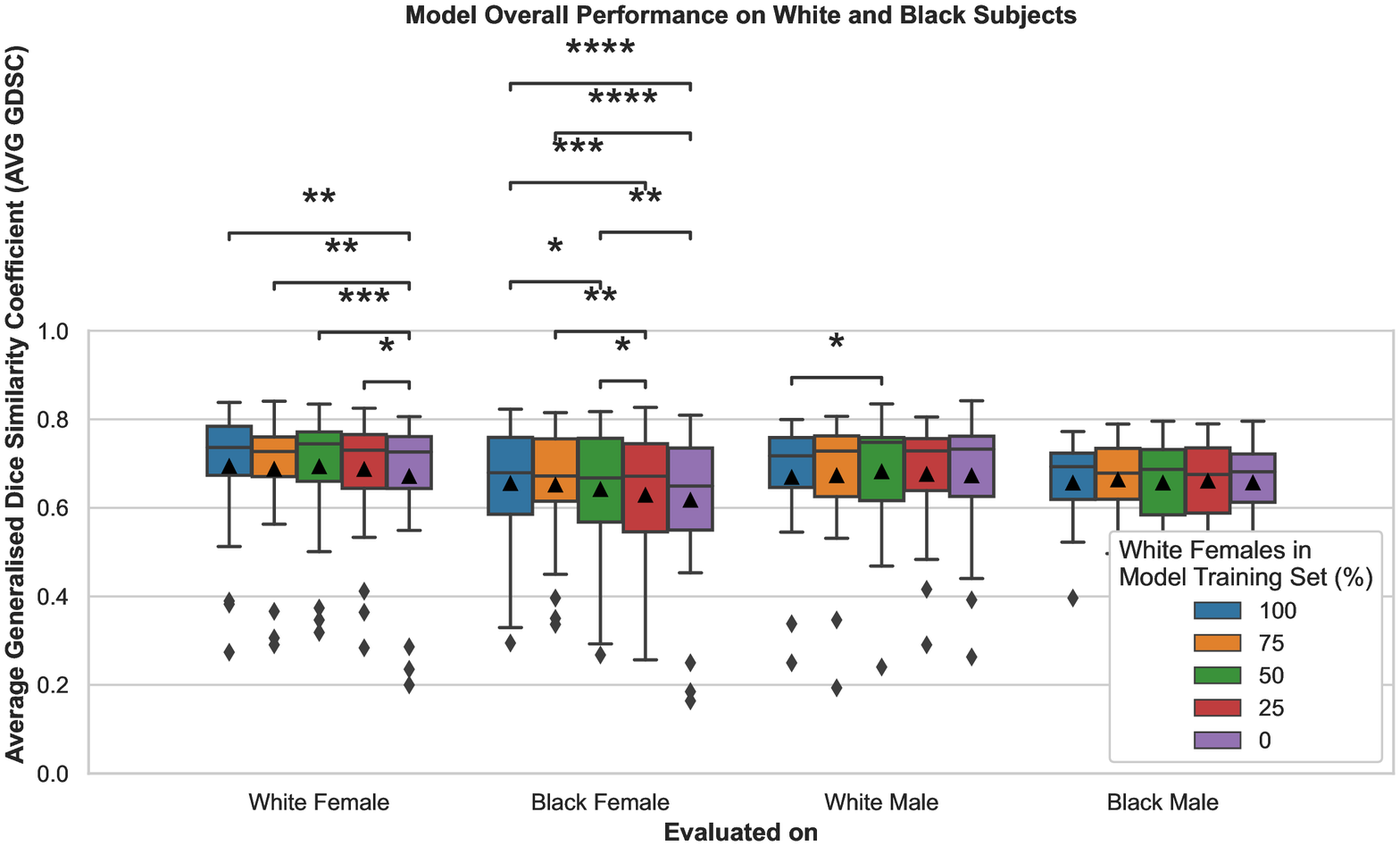}
     \caption{AVG GDSC (excluding background class) on white and black females and males, using models trained on white subjects with different levels of sex imbalance. Significance using Wilcoxon signed rank test \nosignificanceDef.}
     \label{fig:race-bias-overall} 
 \end{figure}

\begin{table}[]
\centering
\begin{tabular}{p{0.4\linewidth}|p{0.6\linewidth}}
\textbf{Comments} &
  \textbf{Labels of the MALPEM atlas segmentation} \\ \hline
\textbf{Each label's Right and Left side is combined into a single label, e.g. \{Right, Left\}-vessel become Both-vessel} &
  Inf-Lat-Vent, Lateral-Ventricle, Thalamus-Proper, vessel, Basal-Forebrain, ACgG-anterior-cingulate-gyrus, AOrG-anterior-orbital-gyrus, AnG-angular-gyrus, Calc-calcarine-cortex, CO-central-operculum, Cun-cuneus, Ent-entorhinal-area, FO-frontal-operculum, FRP-frontal-pole, FuG-fusiform-gyrus, GRe-gyrus-rectus, ITG-inferior-temporal-gyrus, LiG-lingual-gyrus, LOrG-lateral-orbital-gyrus, MCgG-middle-cingulate-gyrus, MFC-medial-frontal-cortex, MFG-middle-frontal-gyrus, MOrG-medial-orbital-gyrus, MPoG-postcentral-gyrus-medial-segment, MPrG-precentral-gyrus-medial-segment, MSFG-superior-frontal-gyrus-medial-segment, MTG-middle-temporal-gyrus, OCP-occipital-pole, OFuG-occipital-fusiform-gyrus, OpIFG-opercular-part-of-the-inferior-frontal-gyrus, OrIFG-orbital-part-of-the-inferior-frontal-gyrus, PCgG-posterior-cingulate-gyrus, PCu-precuneus, PHG-parahippocampal-gyrus, PO-parietal-operculum, PoG-postcentral-gyrus, POrG-posterior-orbital-gyrus, PP-planum-polare, PrG-precentral-gyrus, PT-planum-temporale, SCA-subcallosal-area, SFG-superior-frontal-gyrus, SMC-supplementary-motor-cortex, SMG-supramarginal-gyrus, SPL-superior-parietal-lobule, STG-superior-temporal-gyrus, TMP-temporal-pole, TrIFG-triangular-part-of-the-inferior-frontal-gyrus, TTG-transverse-temporal-gyrus, PIns-posterior-insula, AIns-anterior-insula, MOG-middle-occipital-gyrus, IOG-inferior-occipital-gyrus, SOG-superior-occipital-gyrus \\
  \hline
\textbf{Labels that were removed (mapped to the Background class)} &
  \{Right, Left\}-Cerebral-Exterior, Optic-Chiasm \\ \hline
\multirow{2}{*}{\begin{tabular}[c]{@{}p{\linewidth}@{}}\textbf{Other regions combined into a single label}\end{tabular}} &
  \{Right, Left\}-PIns-posterior-insula, \{Right, Left\}-AIns-anterior-insula \\ \cline{2-2}
 &
  \{Right, Left\}-IOG-inferior-occipital-gyrus, \{Right, Left\}-MOG-middle-occipital-gyrus, \{Right, Left\}-SOG-superior-occipital-gyrus \\
\end{tabular}
\caption{Details of the formation of the atlas we utilised in this study, based on the atlas of the MALPEM segmentations.}
\label{tab:atlas}
\end{table}

\end{document}